\documentclass[conference]{IEEEtran}
\usepackage{times}

\usepackage[numbers]{natbib}

\usepackage{graphicx}
\usepackage{amsmath,amsfonts,bm}
\usepackage{url}
\usepackage{multicol}

\usepackage{subcaption}
\usepackage{booktabs}
\usepackage{makecell}
\usepackage{pifont}   
\usepackage{xcolor}   
\usepackage{listings}
\usepackage{tabularx}
\usepackage{array}    
\usepackage{placeins}
\usepackage{wrapfig}
\usepackage{pifont} 
\usepackage{cuted}

\newcommand{\cmark}{\ding{51}} 
\newcommand{\xmark}{\ding{55}} 
\newcommand{\subsubsubsection}[1]{\paragraph{#1.}}

\newcolumntype{Y}{>{\centering\arraybackslash}X}

\usepackage[bookmarks=true]{hyperref}
\hypersetup{
    pdfauthor={Anonymous},
    pdftitle={Creative Robot Tool Use by Counterfactual Reasoning},
}

\usepackage[most]{tcolorbox}
\tcbuselibrary{listings,breakable}

\newtcblisting{promptbox}{
  listing only,
  breakable,
  enhanced,
  colback=white,
  colframe=black!40,
  boxrule=0.4pt,
  arc=1pt,
  left=4pt,
  right=4pt,
  top=4pt,
  bottom=4pt,
  listing options={
    basicstyle=\ttfamily\footnotesize,
    breaklines=true,
    breakatwhitespace=true,
    columns=fullflexible,
    keepspaces=true,
    showstringspaces=false
  }
}

\title{Creative Robot Tool Use by Counterfactual Reasoning}

\author{
\authorblockN{Mete Tuluhan Akbulut\textsuperscript{*}, Varun Satheesh, Ahmed Jaafar, Alper Ahmetoglu}
\authorblockN{Shane Parr, Aditya Ganeshan, Shivam Vats, George Konidaris}
\authorblockA{Brown University, Providence, RI, USA}
}
\begin{document}

\maketitle

\begin{strip}
\centering
\includegraphics[width=0.98\textwidth]{RSS2026-Counterfactual_reasoning/corl_2025_template/figures/IntroPic.pdf}
\captionof{figure}{An overview of the pipeline. For a given source object and the task definition, a VLM proposes a set of object features that might affect task success. Using the candidate features as axes of variation, a 3D semantic shape editing tool generates counterfactual objects. By carrying out the task in simulation, the robot discovers causal features, which can then be used to identify substitutes for creative robot tool use.}
\label{fig:intro}
\vspace{-0.5em}
\end{strip}


\begingroup
\renewcommand{\thefootnote}{\fnsymbol{footnote}}
\footnotetext[1]{Corresponding author: \texttt{tuluhan\_akbulut@brown.edu}.}
\endgroup

\begin{abstract}
    We propose a causal reasoning framework for creative robot tool use where a suitable tool for a task is correctly identified for use beyond its primary objectives. The proposed framework first discovers the causal relationships between the tool and the task by conducting simulated experiments in a dynamics model. We decouple the causal discovery problem into two complementary components: VLM-based feature suggestion and counterfactual tool generation via targeted geometric and physical feature perturbations. Then, novel objects are classified based on identified causal features, and the tool use skill is transferred via keypoint matching conditioned on the identified causal features. By reconstructing the task in a dynamics model, our approach grounds tool use in the physics of the problem. We illustrate our approach in reaching a distant object with different sticks, scooping candies from a bowl using diverse items, and using different boxes or crates as stepping platforms to retrieve an object from a high shelf. Our baseline comparisons show that identifying causal features and grounding them in physical tool properties leads to more reliable tool selection and stronger skill keypoint transfer. 
    Project page: \url{https://toolanalogies.github.io}.
\end{abstract}

\IEEEpeerreviewmaketitle


\section{Introduction}
Creative tool use is considered an essential trait of intelligence \citep{Call_2013}. Nature provides inspiring examples such as crows that drop stones to increase water level \citep{crows}, and apes that stack crates and use sticks to retrieve a high-hanging banana \citep{MentalityofApes}. Humans can exhibit more complex and creative tool use even at an early age~\citep{Guerin2013}, but emulating the same behavior in robotics remains challenging \citep{Fitzgeraldtoolusage}. Although recent advances improved robot capability to perform complex, contact-rich tasks \citep{Kroemer2021}, creative tool use, which is defined as using tools beyond their intended purpose, is an open problem.

Traditionally, robot tool use has been explored through the lens of affordances, the action possibilities an object offers given an agent’s capabilities \citep{gibson1979ecological, csahin2007afford}. Prior work has focused on learning visual features \citep{Mar2017}, object categories \citep{Sinapov2007}, and keypoints \citep{Turpin-RSS-21} to predict affordances, but these methods often struggle to generalize due to the limited diversity of robot-collected interaction data. Recent approaches leverage large vision-language models (VLMs) trained on web-scale multimodal data to inject commonsense reasoning into tool selection \citep{ahn2022icanisay,driess2023palm}. However, such models often lack grounding in the robot's embodiment and physical environment, leading to failures when proposed tools are physically infeasible or unsuitable for the task.

We propose a causal reasoning framework (Figure \ref{fig:intro}) for creative tool use that enables a robot to identify and repurpose a substitute tool for a task beyond its original intent. Our approach \textbf{generalizes} to novel objects, is \textbf{grounded} in physical constraints, identifies \textbf{causal} features related to task dynamically, and provides \textbf{human-interpretable} justifications for its decisions.
Our key contribution is to integrate the commonsense and generalization capabilities of Vision-Language Models (VLMs) with the physical fidelity of a dynamics model to provide a principled approach to causal feature discovery for tool selection and policy transfer in tool manipulation.
During the \emph{training} phase, our approach assumes that the robot can do the task with a \emph{source} object with a previously acquired skill. It first reconstructs the given task scene in a physics-based simulator using vision-based mesh reconstruction and performs the task with counterfactual objects created by a semantic mesh editor \citep{ganeshan2024parselparameterizedshapeediting} whose physical properties can be edited in the simulator. In doing so, the robot counterfactually discovers the properties of the tool that are both causally relevant to the task and satisfy the physical limitations of the robot. We leverage VLMs to suggest candidate features for the tool, making our approach suitable for open-world domains.
During \emph{deployment}, for each unmodeled \emph{target} object, the robot uses its single-view point cloud to edit the source object along the inferred causal feature dimensions, aligning the edited source object with the target with respect to all causally relevant features. If the edited source object is predicted to succeed through simulation execution, the corresponding target object is suggested for real robot execution. By identifying and reasoning with causally relevant features, our approach is robust to distractors and can readily explain why certain objects were not suitable for the task. 

Beyond tool selection, the inferred causal features also provide a functional basis for policy transfer, because the robot reasons about which object dimensions are relevant to task success, grasp and contact keypoints can be transferred through causal feature alignment rather than through visual correspondence alone. This yields more reliable keypoint transfer under large changes in object shape, appearance, and viewpoint, compared with keypoint matching methods based only on visual similarity, with or without geometric heuristics.
Experiments in three real-world tabletop and mobile manipulation scenarios show that the proposed framework identifies causally relevant, human-interpretable features and finds alternative tools to complete tasks for rigid objects. 

\section{Background and Related Work}
\label{sec:background}

We are interested in one-shot tool manipulation problems. We assume that the robot previously acquired the skill to complete the task with a source tool, and our goal is to identify a novel tool that is a functional substitution for the source tool among many unmodeled objects and to transfer the skill to selected novel tool to complete the task. We studied rigid object interactions in 3 common tool use tasks in literature for comparison: pulling, scooping tasks for table-top manipulation and reaching tasks in mobile manipulation.
We review the related tool use works under the taxonomy of \citet{robotoolsurvey}. Here, we are introducing the most related works. For a more extensive related work section, please see \ref{appendix:related work}.

Part-level affordances that encode the action possibilities are leveraged to enable transfer between intercategorical tools \citep{Austin2015, Schoeler2016, Kroemer2012}. \citet{Fitzgerald2019, Fitzgeraldtoolusage} transferred tooltip pose constraints via human correction trajectories, while \citet{Agostini2015} used an affordance knowledge base for tool substitution in salad-making. We similarly use VLMs as a knowledge base but ground and test the output in simulation.

End-to-end methods predicted actions from visual features for sweeping and hammering \citep{Fang-RSS-18, Finn-RSS-19}. Task-specific keypoint predictors with procedural tool generation improved intercategory use in pushing, reaching, and hammering, using task info as environment keypoints \citep{Qin2020} or rewards \citep{Turpin-RSS-21}. Procedural methods \citep{Qin2020, Turpin-RSS-21, Fang-RSS-18} generated X, L, and T-shaped tools by combining convex parts, requiring 600, 10k, and 18k respectively training tools. By contrast, we use LLM-driven semantic generation, avoiding large-scale training and directly encoding affordances. Combining global and local geometry \citep{liu2024, Qin2021} transfers contact points but still requires demonstrations and ignores physical constraints.

VLMs have recently been applied to reasoning tasks using affordances. Vision work \citep{visionobject, ManipVQA,yu2024} aligns text encodings of tool knowledge with visual task features for tool selection. However, such models lack task dynamics, causal reasoning, and robot interaction data.
In robotics, VLMs have been used for tool selection and policies: \citet{Allen23} trained meta-policies from tool descriptions; \citet{lee2024} used LLMs as symbolic planners; and \citet{car2024} combined high- and low-level planners but limited affordances to grasp prediction. Robotool \citep{robotool}, closest to our work, extracts concepts and plans with privileged knowledge of object layouts, object properties, and explicit task constraints. By contrast, we only reconstruct the source tool, infer or ignore physical properties, and satisfy constraints through simulation. Other approaches generate new tools via VLMs \citep{lin2025robotsmithgenerativerobotictool, gao2025vlmgineer, liu2023learningdesignusetools}, whereas we repurpose everyday objects instead of designing or fabricating novel ones.

Lastly, \citet{chen2025toolasinterfacelearningrobotpolicies} uses object reconstruction like our work to learn tool use policies from human videos and \citet{ZhuPhysical} focused on predefined physical features, which we find essential for tool use.

\section{Counterfactual Reasoning Pipeline for Tool Adaptation}
\label{sec:method}
Consider a TV remote that falls under the sofa where you cannot reach directly. You might look for an object to help retrieve it. Intuitively, you can rule out a book for being too short, a crowbar for being too heavy, or a chair for being too large to fit under the gap. Instead, you might try a rolling pin or a selfie stick. But how do you know which object is suitable for the task? This judgment relies on understanding which physical properties, such as length, weight, or shape, are relevant for the task at hand, and whether a candidate object satisfies those properties. Humans can reason about this effortlessly using prior experience, commonsense knowledge, and an internal physics model. Please refer to \ref{app:motivating_example} to see the motivation in a toy grid environment.

\begin{figure*}[ht]
    \centering
    \includegraphics[width=1\linewidth]{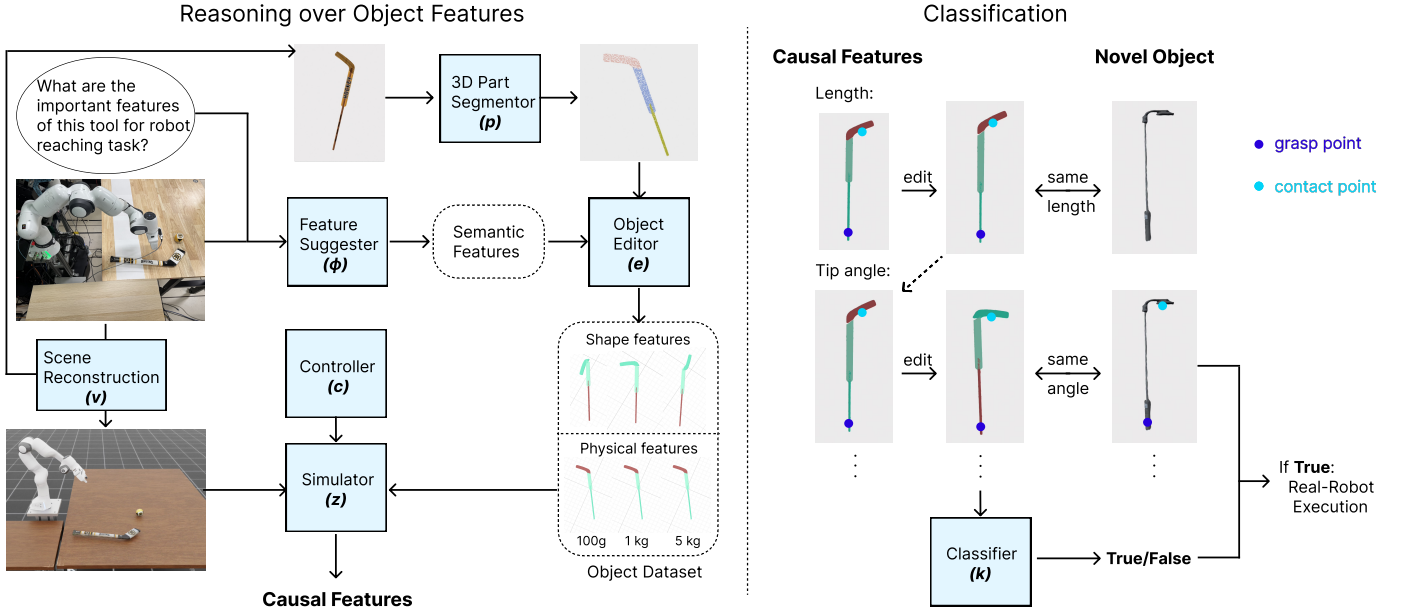}
    \caption{The tool selection pipeline before real-world execution. After finding out the causal features, the source object is morphed to match the dimensions of the target object using the semantic object editor and Chamfer distance as the metric.}
    \label{fig:method}
\end{figure*}

In this work, we show that a similar form of reasoning can be enabled in robots by combining commonsense priors from vision language models (VLMs) with counterfactual reasoning through simulation. We assume that the robot knows how to perform the task with a \textit{source} tool. When placed in a new environment in which the source tool is absent, the robot must identify and use a \textit{novel} (unmodeled) tool for the same task. Without a principled approach, it would spend a lot of time trying objects while violating task and robot constraints. We propose \textbf{ToolAnalogy}, a novel approach that discovers object properties that are causally related to task success by experimenting with counterfactual objects generated with a 3D semantic object editor, and uses these causal features to classify novel objects as substitutes to carry out the task.

\subsection{Problem Formulation}
We formulate our problem by a Markov Decision Process (MDP) $M = \langle S, A, R, T, \gamma \rangle$ where $S$, $A$, $T$, $R$, $\gamma$ denote state space, action space, transition function, reward function, and discount factor, respectively.
We assume that the state can be factored into objects $s = (o_{a}, o_{1}, o_{2}, ..., o_{n})$ where $o_{a}$ denotes the agent and the others denote objects in the environment. The reward, defined by the task description $\mathcal{T}$, is a binary function for task success.
We are interested in tasks in which the robot must use one of the objects as an intermediary tool ($o_{\text{tool}}$) to accomplish the task. 
We model each object using a feature vector consisting of semantic features, which could be exhaustively listed to cover all possible features $F$ and which are all known in common sense knowledge. For example, for a hockey stick, one can say rigid, has an angled head, wooden, thin, etc. and other people could name different things.
However, it is impractical to assume that all can be listed or that this exhaustive list can be used for classification. Therefore, we assume the features are unknown to the robot at the start.
We discover a subset of causally relevant features that are sufficient to satisfy \textbf{the task at hand} $o_{i} = (x_1^i, x_2^i,, ..., x_k^i), X \subset F$. Note that the relevant features of the tool can change between tasks. For pulling, the angled head is causal; for use as a paperweight, it is irrelevant.

\subsection{Method}

Figure \ref{fig:method} outlines our framework.
First, given a task image and a description, a VLM (feature suggester) produces candidate `make or break' tool features that are semantically related to the task at hand. Using these features as axes of variation, a semantic object editor generates a dataset of counterfactual objects. Then, the robot identifies the causal features by experimenting with these objects in simulation by reconstructing the real world task using a real-to-sim-to-real approach. Finally, by constructing a classifier with the causal features, the robot can figure out whether an unseen, unmodeled, tool can be used as a substitute for the task. While simulation can be used as an additional safety check for test objects, our approach incurs no simulation cost at test time once the causal classifier is constructed, avoiding the computational overhead of 3D modeling and physics-based rollouts.


To identify causally relevant features of a tool in the real world, we intervene on its features one at a time to generate modified versions of the tool \citep{Pearl_2009, TabithaCausal}. The robot skill is then executed in simulation on the generated tools to understand the causal effect of interventions on task success. Then, to understand conditional relations, we generate all combinations of causal feature values. Notice that this operation grows the size of the dataset exponentially, so determining causal features first is important. This procedure requires the following functions (annotated in blue boxes in Figure \ref{fig:method}):

\textbf{3D Part segmentor ($p$)}. Here, we use a model that automatically identifies object parts without requiring a text query, as our shape editor, ParSEL, operates on object parts. The part names we are looking for are unknown and vary across different objects. SAMPART3D \citep{yang2024sampart3dsegment3dobjects} satisfies this requirement. It takes a 3D model of the object coming from the scene reconstruction module ($v$), finds its parts with a granularity scale, and names them. This scale is a hyperparameter, and 1.5 works the best for our use case. For the hockey stick in the working example, we get the hockey shaft and the hockey blade as parts.

\textbf{Feature Suggester ($\phi$).} The object feature suggester that gets source object image $I_s$, task description $\mathcal{T}$ which includes implementation details of the controller and returns possible feature values, $\phi(I_s, \mathcal{T})=\{x_1, x_2, \dots, x_n\}$.
We use ChatGPT-5.2 \citep{openai2024chatgpt} as the feature suggester in our experiments.  We leverage VLMs to get semantic features of the source tool because learning causal variables from data is an important open problem \citep{schölkopf2021causalrepresentationlearning}, and VLMs act as a good database for commonsense knowledge to provide candidate features. We run the VLM 10 times and get the top 6 most voted features out of 12. The features of the source stick in Figure~\ref{fig:method} would be shaft length, blade shaft angle, blade length, blade thickness, shaft thickness, and blade width.

\textbf{Object Editor($e$).} The object editor gets a feature $x_i$ and the 3D object model $O$ and returns the edited object model, $e(x_i, O) =O^{*}_{x_i}$.
We use ParSEL \citep{ganeshan2024parselparameterizedshapeediting} to edit 3D objects using semantic shape features. It is based on program synthesis and gives reliable performance in comparison with other data-based methods. It takes object parts and an edit request as input and produces an edit program that can generate shapes with varying scales of that request to create a dataset. For physical features like mass, we edit physical properties in the simulator. See object dataset in Figure \ref{fig:method} where edits for shaft blade angle and mass are illustrated.


\textbf{Scene Reconstruction ($v$).} Vision model to get 3D model of \textbf{only the source tool and goal objects} using images. We are not finding reconstructions of the candidate tools because this could again take a long time for an increasing number of objects. An object model with the same name could be downloaded from open datasets, and its dimensions can be scaled to match the real tool approximately, as an alternative to 3D reconstruction. For reconstruction, we are using ArCode \citep{arcode} and SAM3D \citep{sam3dteam2025} aligned with the recent real-to-sim-to-real pipeline \citet{torne2024reconciling}, which creates the 3D model of the source object $O_{s}$ from its images $I_{s}$, $v(I_s)=O_s$. We assume 3D models of a table for robotic arms and a floor for mobile manipulators. See scene reconstruction in Figure \ref{fig:method}.

    
\textbf{Controller ($c$).}
We represent manipulation skills using task-relevant keypoints, following \citet{liu2024}, which transfers a single demonstration trajectory from a source tool to candidate tools by matching keypoints with DINOv2 and local curvature features. In our setting, the demonstration is likewise specified by keypoints, including the grasp point and the task contact point. For counterfactual tools generated from the source object, keypoint transfer is induced by the causal feature edit itself: as the semantic editor perturbs the relevant geometric or physical feature, keypoints attached to the edited parts are transformed consistently with the object, as shown in Fig.~\ref{fig:method}~(right). For an unmodeled novel object, we first identify the most similar counterfactual tool using Chamfer distance, as described in the Classification section, and then transfer each keypoint to the closest point on the novel object.

\textbf{Dynamics Model ($z$).} We have access to an external dynamics model $z$ where we replicate the real world scene and a robot controller $c$ to achieve the task there. The model takes the edited object meshes $O^*_{x_i}$ and controller $c$, rolls out an episode and returns a boolean to report task success as specified in the task description, $z(\mathcal{T}, O^*_{x_i}, c) = \{0, 1\}$.
We used the simulators IsaacSim \citep{NVIDIAIsaacSim2021} with IsaacLab \citep{IsaacLab} wrapper for table-top manipulation and MuJoCo \citep{Mujoco} for mobile manipulation. Figure \ref{fig:method} shows a scene from IsaacSim.
    
\textbf{Feature Classifier($k$).} The feature classifier checks if the novel (replacement) object has the causally relevant features $x$ found in the reasoning step. It uses RGBD image ${I}_{r}$ of the replacement object and the RGBD images of edited objects in the dataset $\mathcal{O}^*=\{O^*_{x_1}, \dots, O^*_{x_n}\}$: $k(x, I_r, \mathcal{O}^*)$.
This procedure is illustrated in Figure \ref{fig:method}, right. We use the edit functions of each causal feature and make it as similar as possible to the novel object using the Chamfer distance. Figure \ref{fig:method} shows that the length source stick is scaled to match length of the novel object, `the selfie stick'. Then, its tip angle is changed to be the same as the selfie stick. This process continues for all object features and repeated one more time to get the best match. The novel object is classified by checking whether its inferred causal feature values fall within the working ranges established during simulation execution. If the object is classified as suitable, we transfer the keypoints of the skill from the edited source tool to the novel object using Chamfer distance, and then execute the skill on the real robot.
The limitations of this process for hollow objects are discussed in the experiments section (4.4); therefore, we also employ an additional classifier (VLM) to get a second opinion for this type. This time it uses RGBD image $I_{r}$ of the replacement object and the RGBD image of boundary objects in the dataset. Here, boundary objects $O^*_{x_l}$, ${O}^*_{x_h}$ means the objects that successfully worked for the task with the lowest and the highest causal feature ($x$) values: $k(x, I_r, O^*_{x_l}, O^*_{x_h}) = \{0, 1\}$.
        

As our models utilize pre-trained foundational models, no training is necessary in any of the components apart from the skill acquisition procedure.
In case the suggested tool does not satisfy the task, we employ a 2 step solution:
First, we hypothesize that all causal features have not been discovered, and we ask for additional features from the feature suggester and run the pipeline again.
Second, we hypothesize that there are other unaccounted failures like sim-to-real gap, and we continue to try the next suggested tool by the pipeline.

\section{Experimental Results and Discussion}
\label{sec:result}
To evaluate the efficacy of our methodology, we perform experiments in three real-world table-top and mobile manipulation domains involving diverse objects and tool-use scenarios (Figure \ref{fig:domains}). The goal of our experiments is to study (1) the efficacy of our approach in reasoning about novel objects, (2) generalization across diverse tool-use scenarios, and (3) interpretability of the decisions made by our approach. 

\textbf{Baselines.}
We compare our approach with state-of-the-art approaches from the vision and robotics community that leverage VLMs and geometry heuristics for tool selection without requiring any training.  (1) \textbf{ChatGPT-5.2} \citep{openai2024chatgpt} is a strong baseline that has been shown to be capable of performing complicated vision-language reasoning tasks. We provide ChatGPT-5.2 with an image of all the available objects and prompt it to select the most relevant object for the task at hand. We also run one other variation (5.2 t-OBJ) where we add transformed ground truth object meshes to robot frame for source and target objects. (2) \textbf{CoTDet} \citep{visionobject} is a pretrained large vision model trained on the CoCo dataset \citep{lin2014microsoft}, that can predict tool affordances with rationales to support the prediction. We provide an image of all the tools and then sort them based on confidence scores.
(3) \textbf{MAGIC} \citep{liu2024} is a recent robotics approach for generalization of manipulation strategies to novel objects by contact-point matching using \textbf{DinoV2} \citep{oquab2024dinov} features and local curvature analysis. We provide an image of each tool separately and compute confidence scores for keypoint transfer to different objects. We additionally include \textbf{DINOv3} \citep{simeoni2025dinov3} as a strong baseline, reflecting the latest generation of vision backbones for correspondence and representation learning.
(4) \textbf{Groundtruth Object (GT Obj.)} is an ablation baseline where all true target object meshes are reconstructed using \citep{arcode,torne2024reconciling}, and only the classification step is applied to them, i.e., after grasp and contact points detection the objects are imported to the simulator to check their success without added simulator noise.
(5) \textbf{Human}-annotated baseline: We collected responses from 22 participants, each completing a questionnaire under the same conditions as the feature suggester. Specifically, they were presented with the task image, task description, and controller description, along with 12 VLM-generated candidate features, and asked to pick the causally relevant features to the best of their knowledge.

\subsection{Experiment Domains}

\begin{figure*}[t]
    \centering
    \includegraphics[width=\linewidth]{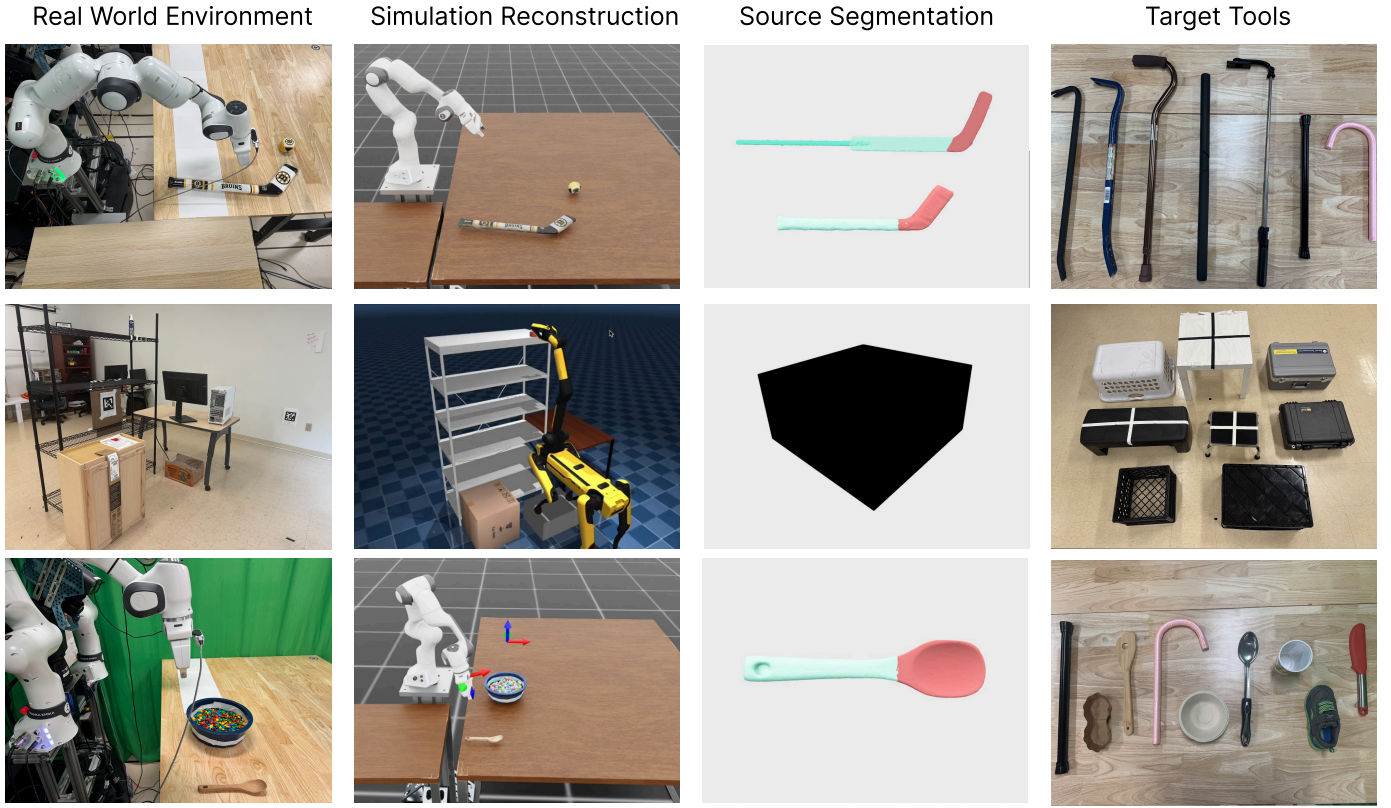}
    \caption{Top---Pulling the ball with a hockey stick. Middle---Reaching an object on the shelf using a platform. Bottom---Scooping candies in the bowl.}
    \label{fig:domains}
\end{figure*}

We compare our approach with baselines in three experiments that challenge the robot to reason about a diverse types of object and object-tool interaction. Note that the objects all have different primary functionalities in daily tasks. All objects are placed in 10 cm for pull and 5 cm for other tasks ranges in the x and y axes, and all experiments are run for 10 seeds.

\textbf{Table-top Pulling} requires the Franka Emika Panda robot to grasp a tool to pull an object located outside of its workspace (Figure \ref{fig:domains}, top). A suitable tool must meet both the task requirements like length and angled tip and also satisfy the robot's physical limitations, such as the payload. We assume that the robot has the skill to pull the toy puck with the toy hockey stick, and when it is in a new scenario where it needs to pick a new tool among the blue crowbar, selfie stick, walking cane, shepherd cane, black crowbar, yoga stick, curtain hanger. The task is satisfied if the puck is in reaching distance. Here, we run two versions of the task where we use the toy hockey stick model after reconstruction and just download and use a real hockey stick model as source objects. 

The controller is defined using two keypoints: handle point and tip point. The robot grasps the tool at handle point, and brings the tip point behind the object and pulls the tool toward the initial grasp. Mass of the tool is determined by torque values of the robot after lifting it. The execution is stopped if the mass is over the working range found using the simulator.

\textbf{Table-top Scooping} requires the Franka Emika Panda robot to grasp a tool to dip it inside a bowl and scoop to obtain small candies (Figure \ref{fig:domains}, bottom). A suitable tool must meet both the task requirements like handle thickness and head curvature and also satisfy the robot's physical limitations, such as the payload. We assume that the robot has the skill to scoop some candies with a wooden spoon, and  when it is in a new scenario where it needs to pick a new tool among black curtain rod, cartboard egg bite tray, wooden spatula, pink shepherd cane, cartboard bowl, metal serving spoon, plastic cup, toy shoe, and red scraper. The task is satisfied if the robot lifts any candy with the help of the tool.

The controller is implemented using 2 keypoints, handle and contact point. The robot grasps the object by the handle point, carries the tool over the bowl, then rotates the tool such that the tip point faces downwards, dips the tool into the bowl until the tip point is in contact with the candy, and lifts the tool back up while rotating it back to horizontal.

\textbf{Quadruped Reaching} challenges a Spot robot with an arm to retrieve an object from a shelf that is beyond its reach (Figure \ref{fig:domains}, middle). Unlike the pulling task, this requires the robot to use an object as a stepping tool to increase its reach and retrieve the object. A suitable tool must both be able to extend the robot's reach and also to fit between 2 obstacles to be placed. We used MuJoCo \citep{Mujoco} as the simulator. The candidate tools shown on middle in Figure \ref{fig:domains} are a laundry basket, a small and a large milk crate, a stepping stool, a black gripper box, a gray gripper box, an IKEA table, and an aerobic step. The task is satisfied if the robot can reach to the object on the top shelf.

\textit{The controller in the simulator}: We do not have access to the walking behavior that Spot has in the real-world, and as such, we spawned it on the platform in the simulator to resemble the behavior\footnote{For the real-world walking behavior, we contacted RAI \citep{RAI} but could not obtain the controller in simulation.}. We control the arm of Spot using its kinematics model. Following \cite{torne2024reconciling}, we downloaded similar items for the shelf and the goal object from the internet.


\textit{The controller in the real world}:
We utilize the Boston Dynamics Spot SDK to control the robot. The skill is written as moving forward by a specified distance towards the goal. During this forward trajectory, Spot autonomously detects and steps onto an obstacle, in this case, a stepping tool, without requiring explicit modification by the built-in controller, which again is not exposed to outside. Following this movement, we command Spot's onboard arm to reach to the shelf by pose commands. Note that there is no keypoint for the stepping behavior, so keypoint baselines are removed.

\begin{table*}[ht]
  \caption{Task performance, human supervision cost, and interpretability across methods}
  \label{tab:merged_results}
  \centering
  {\footnotesize
  \setlength{\tabcolsep}{2pt}
  \renewcommand{\arraystretch}{1.12}

  \begin{tabularx}{\textwidth}{l *{8}{Y}}
    \toprule
        & Ours
        & GPT-5.2
        & GPT-5.2 (t-OBJ)
        & DINOv3
        & MAGIC
        & CoTDet
        & GT Obj. \\
    \midrule

    \multicolumn{8}{c}{\textbf{Task Performance}} \\
    \midrule

    Table-top Pulling   & 90\% $\pm$ 20\%  & 20\% $\pm$ 24.5\% & 15\% $\pm$ 23\% & 50\% & 50\% & 50\% & 100\%\\
    Table-top Scooping  & 40\% $\pm$ 20\%   & 0\% $\pm$ 0\% & 0\% $\pm$ 0\% & 0\%  & 0\%  & 0\% & 50\% \\
    Quadruped Reaching  & 66.7\% $\pm$ 25.8\% & 0\% $\pm$ 0\% & 0\% $\pm$ 0\% & N/A & N/A & 66.7\% & 100\%  \\

    \midrule
    \multicolumn{8}{c}{\textbf{Human Supervision and Interpretability}} \\
    \midrule

    Human supervision   & \xmark & \xmark & \xmark & \xmark & \xmark & \xmark & \cmark  \\
    Interpretability    & \cmark & \cmark & \cmark & \xmark & \xmark & \xmark & \xmark  \\

    \bottomrule
  \end{tabularx}%
  }
\end{table*}

\subsection{Comparison with baselines}

\begin{figure*}[ht]
    \centering
    \includegraphics[width=\linewidth]{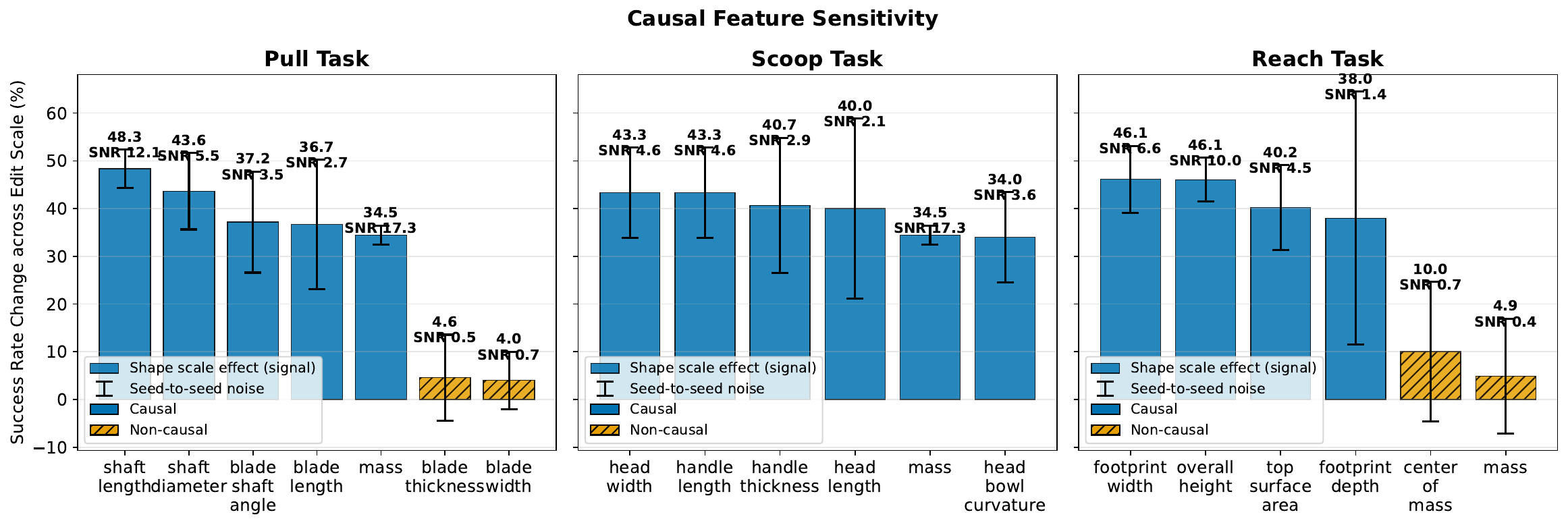}
    \caption{Perturbations of identified causal features (blue) yield larger success-rate changes than non-causal features (yellow) across pulling, scooping, and reaching; error bars show seed variability for 10 seeds.}
    \label{fig:feature_sensitivity}
\end{figure*}



    

Table~\ref{tab:merged_results} shows the success of the methods in predicting suitable tools among all objects in the environment for 10 seeds. Ground-truth values for all target tools were obtained by executing the skill in the real robot setup. We report how many suitable tools were found within the top-$n$ guesses, where $n$ is the number of suitable tools, because DINOv3, MAGIC, and CoTDet only generate confidence scores that can be converted into preference lists. Since these baselines are deterministic, we do not report variance for them. ToolAnalogies achieves higher accuracy than all baselines, except for the ablation with ground-truth object models.

Across experiments, the baselines reveal complementary failure modes. Vision-only methods, including DINOv3 features and VLM image prompts, capture coarse shape similarity but do not reliably account for task-specific physical constraints, often ranking visually plausible but physically unsuitable tools, such as payload-inappropriate crowbars, too highly. Adding local geometric cues through MAGIC improves sensitivity to contact geometry, but projected local descriptors are insufficient for functional properties that depend on global 3D structure, such as containment, concavity, length, or mass. Similarly, providing target-object geometry to GPT-5.2 does not resolve these failures, suggesting that richer geometric context alone is insufficient without task-level physical reasoning. CoTDet performs well when familiar affordance categories align with the task, as in quadruped reaching, but remains weak for novel object use and environment-specific constraints. These results indicate that successful creative tool use requires reasoning over the causal features that mediate task success, rather than ranking objects by visual similarity, object category, or local geometry alone. ToolAnalogies addresses this gap by generating counterfactual variations of the source tool along candidate feature dimensions and testing their effects in simulation, thereby identifying which geometric and physical properties are causally relevant under the robot's embodiment and environment constraints.

The GT Obj. ablation serves as a privileged upper bound rather than a deployable baseline, since it assumes reconstructed target-object models and simulator execution for each candidate tool. Its strong performance confirms the value of physical testing, but also exposes the cost of requiring full target-object reconstruction. ToolAnalogies achieves the strongest performance among non-privileged methods while avoiding exhaustive target-object modeling, which is important in open-world settings where multi-view scanning and simulation-ready model preparation for every candidate object are costly and often impractical.

Beyond task performance, Table~\ref{tab:merged_results} also highlights differences in supervision and interpretability. ToolAnalogies uses the source object model to generate counterfactual variations, but evaluates target objects through inferred causal features rather than full reconstruction. Its interpretability comes from causal feature discovery: VLM-proposed candidate features are grounded through object edits and tested in simulation, allowing selected and rejected tools to be explained by task-relevant geometric and physical properties. VLM baselines can also provide textual explanations, but these explanations are not physically grounded or tested in the robot's environment. Failure cases of our method are detailed in Section~\ref{sec:failures}.

\begin{figure*}[ht]
    \centering
    \includegraphics[width=\linewidth]{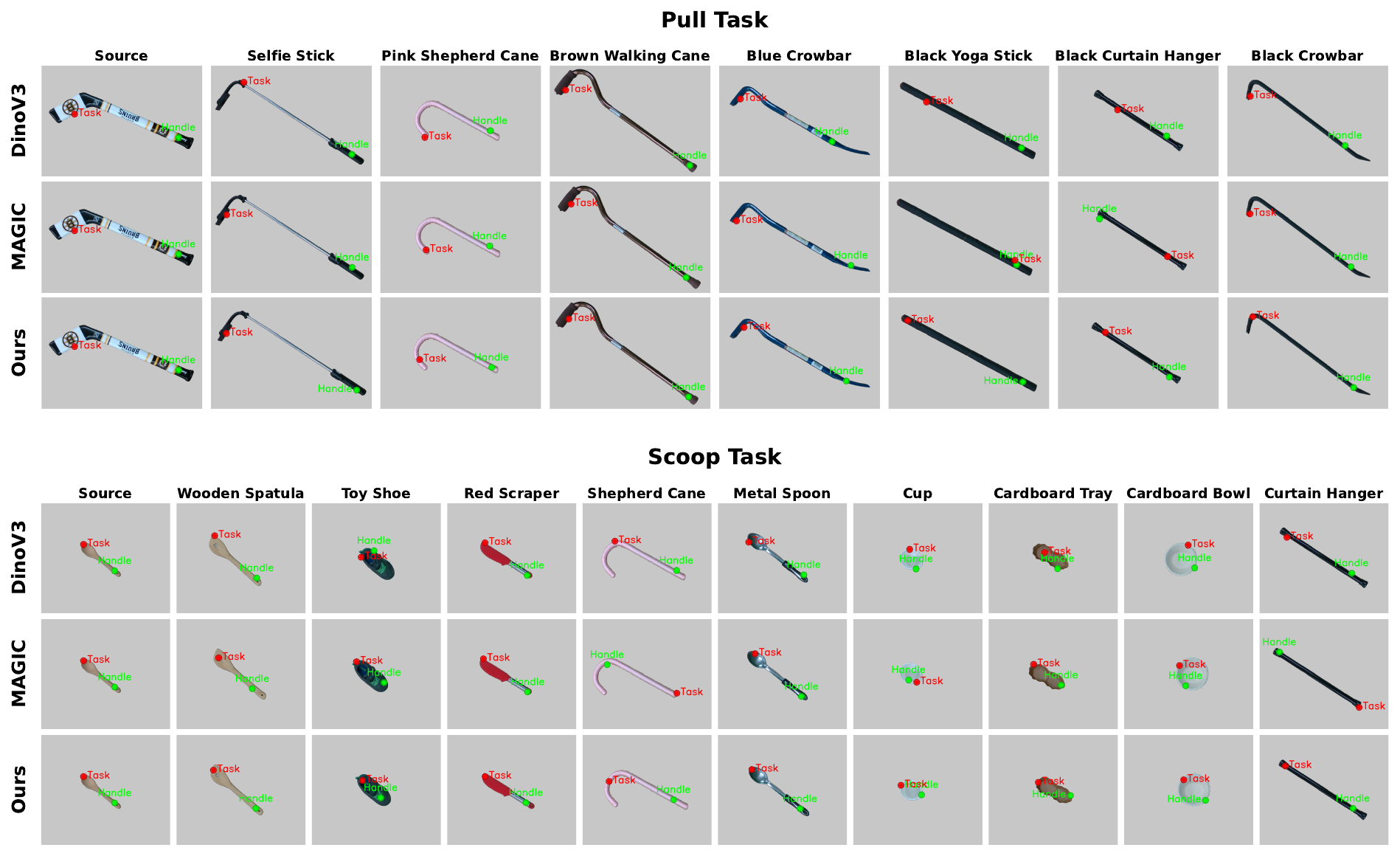}
    \caption{Keypoint transfer.}
    \label{fig:keypoint_transfer}
\end{figure*}

\subsection{Causal Feature Analysis and Human Alignment}

Figure~\ref{fig:feature_sensitivity} analyzes whether the features identified by our pipeline have a measurable causal effect on task success. For each task, we perturb one feature at a time across its edit scale while holding the remaining features fixed, and report the induced change in success rate relative to seed-to-seed variability. Across pulling, scooping, and reaching, the identified causal features produce substantially larger success rate changes than non-causal features. In the pulling task, shaft length, shaft diameter, blade-shaft angle, blade length, and mass have strong effects, whereas blade thickness and blade width remain close to the noise level. In scooping, all retained features affect success, including head geometry, handle geometry, mass, and bowl curvature. In reaching, footprint width, overall height, top surface area, and footprint depth dominate, while center of mass and mass have weak effects. These results support the central claim that tool suitability is governed by a small set of task dependent causal features rather than by global visual similarity alone.

\begin{wraptable}{r}{5.9cm}
\centering
\caption{Causality analysis}
\label{table:causality}
\scalebox{0.68}{
\begin{tabular}{|c|c|c|}
\hline
 & \makecell{Human-alignment} & \makecell{Useful features\\ for target selection} \\ \hline
Pulling & 83\% & 5/5 \\  \hline
Reaching & 66\% & 3/4 \\  \hline
Scooping & 86\% & 3/6 \\  \hline
\end{tabular}
}
\end{wraptable}

Table~\ref{table:causality} reports the contribution of the inferred causal features to task success, and the similarity of inferred causal features to the causal features suggested by the human study. We labeled a feature 'causal' for human baseline if more than 50\% of participants agree. We report the percentage of overlapping features with our pipeline. See Appendix \ref{appendix:human_survey}.

Across all tasks, our approach identifies causally related features that are sufficient to classify suitable tools. However, this does not imply that it recovers every causal feature selected by humans. In the survey, participants sometimes labeled synonymous or highly overlapping features as causal, whereas our pipeline tests features incrementally after the initial VLM-suggested candidates and is therefore less likely to recover redundant variants of the same property. Moreover, because target objects are not known during feature generation, both humans and our method may identify features that are plausible for the task but not necessary for distinguishing the specific target objects in our experiments. Finally, some human-selected features, such as friction, are not currently supported because estimating them reliably in the real world requires additional interaction.

\subsection{Skill Transfer via Keypoints}

Figure~\ref{fig:keypoint_transfer} shows a qualitative comparison of keypoint transfer for pulling and scooping. The red point marks the task/contact keypoint and the green point marks the handle/grasp keypoint. Our approach achieves qualitatively better keypoint transfer than visual matching baselines, including methods with geometric heuristics. DINOv3 underperforms because it transfers keypoints mainly through visual similarity, which leads to failures on objects such as the black yoga stick and selfie stick in pulling, and the toy shoe and cardboard bowl in scooping. MAGIC improves performance with a fixed curvature heuristic, but fails to generalize to objects with low curvature variation, as seen for the black yoga stick and black curtain hanger in pulling, and the cup and cardboard bowl in scooping. Rather than matching keypoints solely based on visual similarity or fixed local geometry, our method transfers them through the causal feature dimensions identified for the task. This leads to more functionally meaningful correspondences: the grasp point and task contact point are placed on object regions that support the intended interaction, even under substantial variation in object geometry and appearance. Our method performs poorly only on the iron crowbar, where suboptimal optimization leads to an inaccurate keypoint assignment; however, this object is classified as unsuitable due to its weight, so the keypoint transfer error does not affect downstream task success. These results suggest that dynamically grounded causal features provide a stronger basis for policy transfer than visual similarity or predefined geometric correspondence alone.

\subsection{Classification Metrics}
\label{sec:classification_metrics}

\begin{figure}[th]
    \centering
    \begin{subfigure}[t]{0.32\columnwidth}
        \centering
        \includegraphics[width=\linewidth,height=2.5cm]{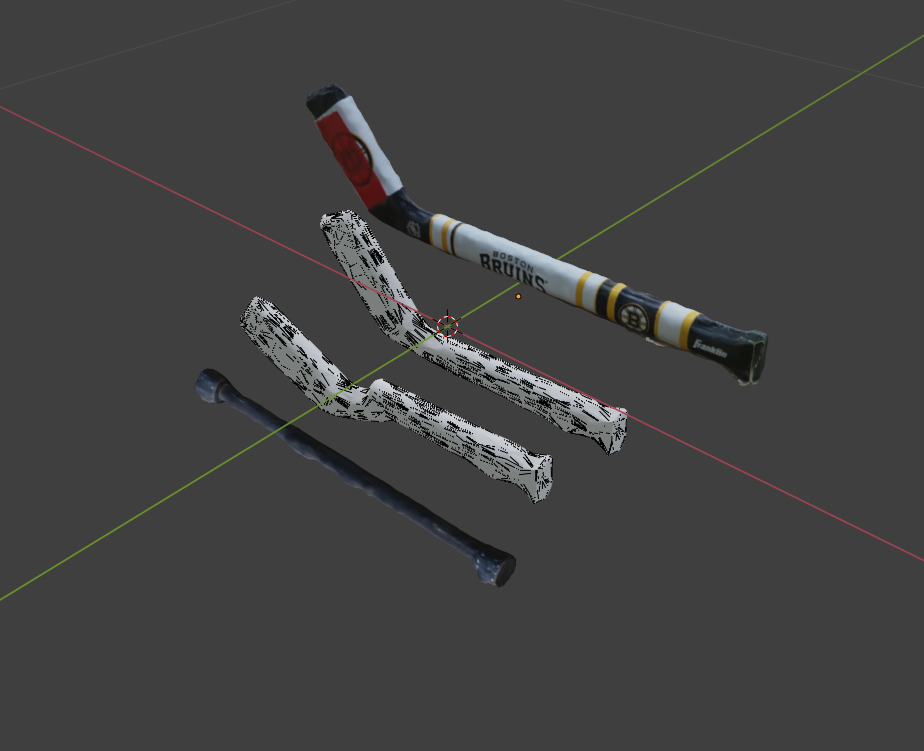}
        \caption{Tool matching success}
    \end{subfigure}
    \hfill
    \begin{subfigure}[t]{0.32\columnwidth}
        \centering
        \includegraphics[width=\linewidth,height=2.5cm]{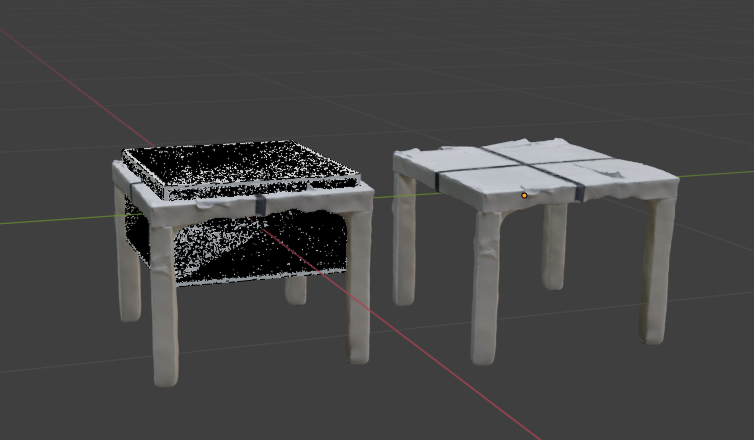}
        \caption{Tool matching failure}
    \end{subfigure}
    \hfill
    \begin{subfigure}[t]{0.32\columnwidth}
        \centering
        \includegraphics[width=\linewidth,height=2.5cm]{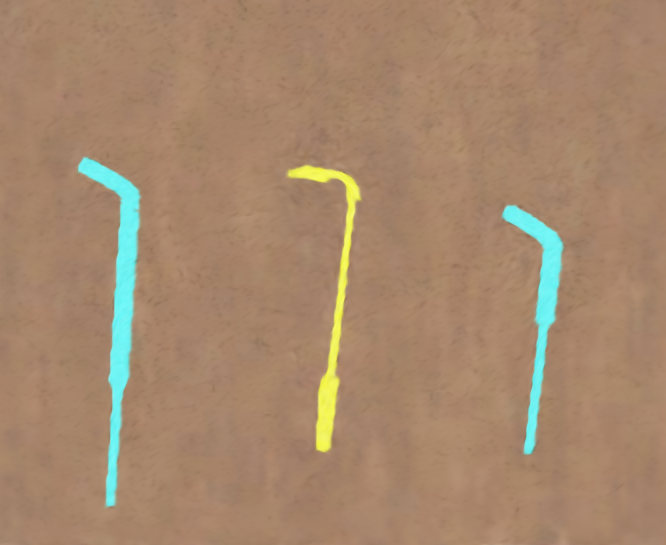}
        \caption{VLM classification}
    \end{subfigure}
    \caption{Classification methods}
    \label{fig:classification}
\end{figure}

\begin{figure*}[ht]
    \centering
    \includegraphics[width=\linewidth]{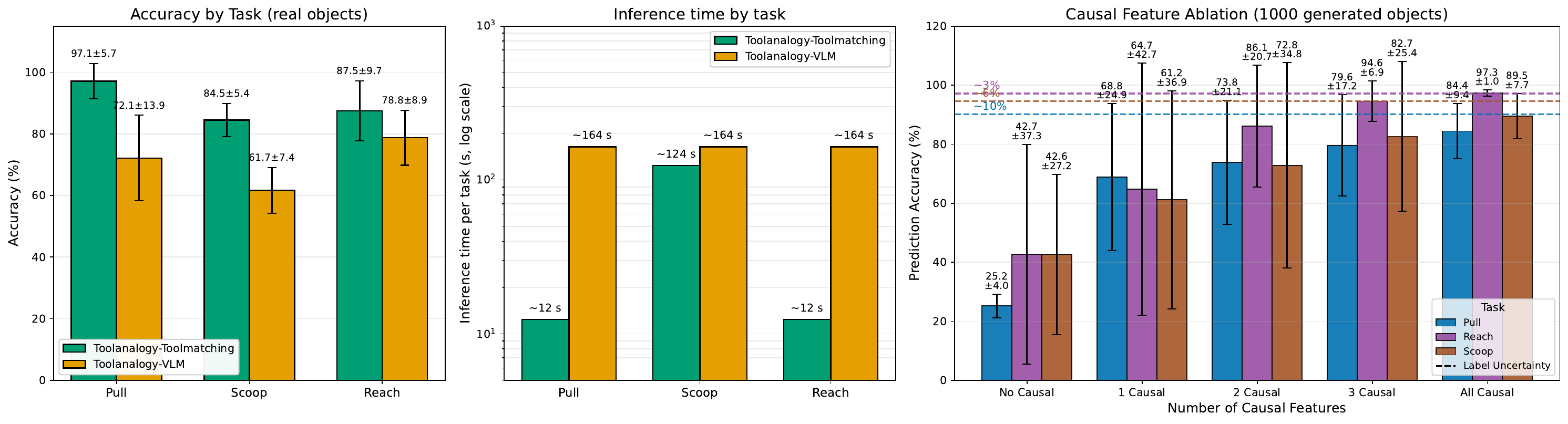}
    \caption{Ablation studies}
    \label{fig:ablation_studies}
\end{figure*}

In the Figure~\ref{fig:classification}, we explain the different classification methods mentioned throughout this paper. Image a illustrates our classification approach. We transform our source object (hockey stick) through two successive operations designed to approximate the geometry of a curtain rod. In contrast, Image b presents a failure case of the proposed classification approach. This failure arises for a hollow table mesh: fitting the source box to the target includes substantial empty space, and the resulting Chamfer distance match prevents the source object edit from recovering the correct table height. Image c illustrates the alternative approach for such failures. We render the partial pointclouds of the target tool and the two boundary source tools that bracket the continuous operational range for the causal feature under consideration. Here, the pointclouds of the shortest and longest suitable sticks are shown in blue. Any stick whose length falls between them is also suitable. The rendered image is then fed to the VLM to determine whether the target object's length lies within the specified range. This differs from the VLM baseline, since our method first identifies the relevant causal features and their working ranges, rather than directly asking the VLM to classify object suitability.

\subsection{Classifier Comparison and Causal Feature Ablation with Generated Shapes}

Figure~\ref{fig:ablation_studies} compares the two classification strategies and analyzes the role of causal features. 
\textbf{Left:} We report the complete object-level classification accuracy on real target objects. This differs from Table~\ref{tab:merged_results}, where a method is counted as successful if it can identify suitable tools within the top $n$ ranked candidates, with $n$ equal to the number of suitable tools. The VLM-based classifier performs worse because it is not physically grounded in the task constraints and often fails to distinguish small but functionally important shape differences, such as small changes in angle, length, or thickness. For the reaching task, tool matching uses the VLM-based decision only for hollow objects, due to the failure mode discussed in Section~\ref{sec:classification_metrics}.
\textbf{Middle:} We report the inference time of the two classifiers. The VLM classifier is evaluated with the same 10-vote mechanism and takes approximately 134 seconds per object. The inference time of tool matching depends on the number of inferred causal features: it takes approximately 12 seconds for pulling and scooping, but approximately 120 seconds for reaching, indicating combinatorial growth with the number of features. This motivates identifying causal features before classification, since evaluating all causal and non-causal features would further increase runtime. These times are measured without code optimization or parallelization, so absolute runtimes could be reduced. More importantly, the results show why restricting classification to inferred causal features is essential: it avoids unnecessary evaluation over non-causal dimensions and keeps the search space tractable.
\textbf{Right:} We evaluate causal feature ablations on 1000 generated objects, produced by applying both causal and non-causal feature edits. The horizontal dashed lines show the average pairwise disagreement between simulator seeds when the generated shapes are evaluated in simulation, which reflects label uncertainty due to simulator noise, and therefore sets an upper bound for the achievable prediction accuracy. As expected, the accuracy of the prediction improves as more inferred causal features are included. The remaining gap between the dashed line and the bars using all causal features reflects the limitation of the Chamfer distance optimization. This gap is small for the reaching task, where the edits are simple box transformations, but larger for the spoon and stick tasks. The "No Causal" setting varies only non-causal features; since these edits preserve the functional properties of an initially successful object, the generated set is dominated by successful simulator labels. The corresponding accuracy therefore reflects the label distribution of the generated test set.



\begin{table*}[!ht]
\centering
\caption{Failure analysis of the proposed pipeline and its mitigation, bold points highlight modular design of the pipeline}
\label{tab:failure_analysis}
\small
\setlength{\tabcolsep}{4pt}
\renewcommand{\arraystretch}{1.12}
\begin{tabularx}{\textwidth}{@{} l r Y Y @{}}
\toprule
\textbf{Failure source} &
\textbf{Metric} &
\textbf{Example} &
\textbf{Mitigation} \\
\midrule
Feature suggester &
4\% causal features missed &
Mass initially omitted &
Iterative feature expansion \\

Object editor &
14\% proposed features uneditable &
Unsupported geometric or physical edits &
Grammar-constrained prompts; \textbf{improved editor} \\

Part segmentation &
$\sim$66\% success &
Incorrect part decomposition &
Replaceable segmentation module \\

Chamfer classifier &
5-6\% drop vs. ground truth &
Partial point-cloud thickness ambiguity &
Pointcloud completion \\

Sim-to-real / controller &
2/24 failures &
Toy shoe; metal spoon &
Try next tool; \textbf{better world model} \\
\bottomrule
\end{tabularx}
\end{table*}


\subsection{Failure Cases}
\label{sec:failures}

We analyze the failure modes across the pipeline components and quantify their impact.

\textbf{Feature suggestion.}
The VLM-based feature suggester may omit relevant causal features in early iterations. Using 10 stochastic VLM inferences with majority voting, we observe a $\sim$4\% miss rate (false negatives only), while effectively filtering false positives. Iterative querying mitigates this issue in many cases (e.g., initially missing tool mass), but if a causal feature is never proposed, it cannot be recovered by downstream modules. This defines a bound on achievable performance.

\textbf{Object editing.}
The semantic editor is constrained by the expressivity of ParSEL and the underlying dynamics model. Empirically, $\sim$14\% of proposed features cannot be realized due to unsupported geometric or physical transformations. We partially address this by conditioning prompts on the editor grammar to enforce feasible edits. Remaining failures are unrecoverable within the current framework, though the modular design allows future replacement with more expressive editors.

\textbf{Segmentation.}
Part-level reasoning relies on accurate object decomposition. We employ the off-the-shelf SamPart3D module, which achieves $\sim$66\% success in our setting. While this introduces occasional errors in part identification that may affect feature grounding, the performance is sufficient to support reliable end-to-end behavior in our experiments. Importantly, our framework is modular, and the segmentation component can be readily replaced with more advanced methods as they become available, enabling improved robustness in more complex scenes.

\textbf{Point cloud classification.}
Our classifier relies on single-view point clouds and Chamfer distance, which limits sensitivity to volumetric properties such as thickness. This explains the observed $\sim$5-6\% performance gap in push and scoop task relative to ground truth in Figure \ref{fig:ablation_studies}. We partially mitigate this by augmenting point clouds with symmetry-based completion, but multi-view reconstruction or learned geometric representations would be required for robust handling of such features.

\textbf{Sim-to-real and control.}
Discrepancies between simulated and real dynamics lead to 2/24 misclassifications, primarily due to unmodeled physical properties (e.g., deformability of a toy shoe, mass distribution of a metal spoon). As our framework is designed for one-shot deployment without online adaptation, we handle these cases by proceeding to the next candidate object. Incorporating system identification or adaptive control (e.g., RL-based refinement) is a promising direction but beyond the scope of this work.

\section{Conclusion}
\label{sec:conclusion}

We introduced a causal reasoning framework for creative tool use that identifies and repurposes novel objects as tool substitutes in diverse scenarios. During training, the robot performs vision-based mesh reconstruction to generate a physical reconstruction of the scene in a physics-based simulator. The simulator serves as a causal reasoning engine, where  we  systematically intervene on semantic properties of tools and evaluate the effect on task success to identify the causally relevant features. These features are then used by a classifier, which selects the most suitable substitute tool from the available objects.
Our approach generalizes to novel objects by leveraging the commonsense reasoning of VLMs, is grounded in physics by our use of physics-based simulation and provides interpretable justification by virtue of causal reasoning. Real-world experiments in three table-top and mobile manipulation domains show that it outperforms baselines in creative robot tool use, significantly enhancing the open world manipulation capabilities of robots. 


\section{Acknowledgments}
This research is partially funded by the ONR under the REPRISM MURI N000142412603.

\bibliographystyle{plainnat}
\bibliography{RSS2026-Counterfactual_reasoning/corl_2025_template/references}

\appendices
\section{Supplementary Material}
\label{sec:appendix}

\subsection{A Motivating Example}
\label{app:motivating_example}
\begin{figure}[ht]
    \centering
    \includegraphics[width=0.8\linewidth]{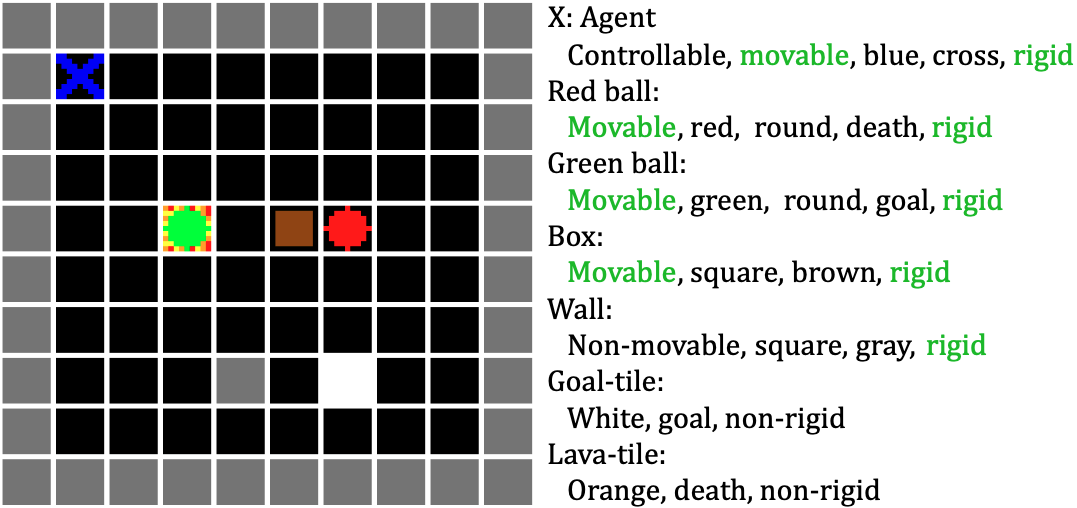}
    \caption{Toy grid‑world used to illustrate our problem setting in an idealized scenario where object features are pre-listed and directly editable unlike real-world: the blue X agent navigates walls and interacts with objects whose attributes (movable, goal, death, etc.) are listed on the right.}
    \label{fig:grid}
\end{figure}

Consider the toy grid environment in Figure \ref{fig:grid} with objects listed on the right. The task is to move the green ball to the goal while avoiding the lava tile.
When the green ball is on lava, the agent must first free the ball by pushing the brown box---an intermediary object---onto the lava tile, and then move it to the goal. In a new environment in which there are no boxes, the agent must find another object that can functionally substitute for the box. Trying each object as the replacement might take an arbitrarily long time in environments with many objects, especially when the evaluation is costly (e.g., the use of a dynamics model) and poses a safety concern (e.g., stepping on the lava tile).
It would be reasonable to pick objects based on their features if only the agent knew which of those are causally relevant.
However, the agent must conduct interventional experiments \citep{Pearl_2009} to find out such features.
For instance, by changing the features of the box---by an intervention---and observing its effects, the agent can identify that the replacement for the box should be rigid and movable while its color is irrelevant, and figure out that the red ball can be used instead. While changing the object properties in game-like environments is possible, doing so in real world scenarios is non-trivial. In this paper, we demonstrate that a similar approach can address real-world tool manipulation problems in which the features cannot be listed as in this example and cannot be easily manipulated in the real world.

\subsection{Extended Related Work}
\label{appendix:related work}

Tool use has traditionally been studied under the topic of affordances \citep{gibson1979ecological}. In push and pull tasks with various shaped sticks, affordances are found by clustering object-effect categories using predefined actions or motor babbling \citep{Sinapov2008, Stoytchev2005, Tikhanoff2013}. To achieve tool transfer beyond categorization, object features are learned with virtual tools in simulation \citep{Mar2017, Nishide2012, TAKAHASHI2017115, TEKDEN2024}. \citet{Gonçalves2014} investigated the role of predefined shape descriptors with visual feature learning. These methods haven't been scalable to different tools due to the limitation of collecting robot experience at the order of the complexity of tool use tasks.

Keypoint representation is another widely used approach to select suitable tools and manipulation policies \citep{kpam}. \citet{tee2022framework} transfers keypoints on robot limbs to tools to attribute limb functionality. \citet{mao23a} used contact points on objects and task and motion planning \citep{Tamp} in balancing tasks in addition to push and pull tasks. 
\citet{kPam2} coupled key points with a feedback controller for wiping and peg insertion. These works addressed tool use problems where tools look similar and arguably belong to the same category.

To accomplish transfer between intercategorical tools, part-level affordances are sought \citep{Austin2015, Schoeler2016, Kroemer2012}. \citet{Fitzgerald2019, Fitzgeraldtoolusage} used human correction trajectories to transfer tooltip's pose constraints for hooking, sweeping, and hammering tasks. In \citet{Agostini2015}, an affordance knowledge base is used for tool substitution to satisfy plans in salad-making tasks. We are using VLMs similarly as a vast knowledge base. However, we are checking and grounding suggestions by reconstructing the task in simulation.

End-to-end methods were also used to predict actions using visual features to use different tools in sweeping and hammering tasks \citep{Fang-RSS-18, Finn-RSS-19}. In addition, task-specific key point predictors, along with procedural tool generation, have shown improvements in intercategory tool use in pushing, reaching, and hammering tasks. Task information was provided as environment keypoints \citep{Qin2020} or as a reward function to the network \citep{Turpin-RSS-21}. In 
\citet{Qin2020, Turpin-RSS-21, Fang-RSS-18}, tools of the X, L and T shape were generated procedurally by combining convex parts. Instead, we use semantic information from LLMs for tool generation so that generated tools are directly related to semantic features, and we do not need any additional training to learn affordance features. In addition, these procedural generation methods \citep{Qin2020, Turpin-RSS-21, Fang-RSS-18} used 600, 10000, and 18000 tools, respectively, for training compared to our 120. In \citet{liu2024, Qin2021}, global and local geometric features are used together to transfer contact points and for tool selection. Although this approach allows transfer between tools that have geometrically similar parts (it requires demonstrations, and) it does not reason about physical features and environment constraints.

In \citet{Abelha2017, Gajewski2019}, pointclouds are processed to characterize tools, grasp and contact segments of the tool are found flexibly based on the task. \citet{Abelha2017} used CAD-models from web to populate a dataset of 5000 tools to test in simulation, and a Gaussian Process is fitted to their affordance score, to classify new objects later for tool selection. This approach was tested in cutting, lifting, hammering, and rolling tasks. Using web models is an alternative to tool generation, but again, our approach does need additional training with thousands of tools to find affordances due to semantically meaningful generation. \citet{ZhuPhysical} considers predefined physical features for the task differently from previous work, which we show is essential for robot tool use.

Recently, Visual Language Models have been used to benefit from web-scale data for tool affordance. The line of work \citet{visionobject, ManipVQA,yu2024} in the vision community uses the similarity between the text encoding of general knowledge of tools queried from VLMs, and the visual encoding of task images to propose and segment a suitable tool for the task. Although we also consider VLMs to be great resources for general information on prior experiences, using them alone is problematic in robotics. By definition, they lack interaction data for the current task, therefore knowledge of task dynamics, causal relationships, and the action capabilities of the agent.

In robotics, VLMs have gained attention in tool selection and manipulation policy as well. \citet{Allen23} uses language descriptions of tools as affordance features to train a meta-policies for sweeping, hammering, pushing, and lifting tasks. In \citet{lee2024}, Large Language Models(LLMs) are used as a high-level symbolic planner for bimanual pull and push tasks. In \citet{car2024}, high-level planner was combined with a low-level planner and vision modules to use tools with LLMs, but tool selection is not addressed, and affordance prediction was limited with grasp point prediction. Robotool \citep{robotool}, closest to our work, used LLMs to analyze the problem to extract key concepts, create plans selects a tool, and execute parametrized skills for reaching, grasping, and pressing tasks. The framework uses privileged task information such as the layout of objects, the positions, sizes, and physical properties of objects, grasp points on objects, as well as robot and environment constraints. On the other hand, we get the accurate reconstruction of only the source tool, physical properties are either inferred through interaction or not used, and satisfying robot and task constraints is learned through a simulator without any explicit definition. Other methods like \citet{lin2025robotsmithgenerativerobotictool, gao2025vlmgineer} address tool design problem \citep{liu2023learningdesignusetools} via VLMs. Here, we are trying to repurpose an everyday object to satisfy the task instead of designing and printing 3D shapes.

\subsection{Computer specifications and VLM calls}

The computer that we used to run our pipeline has Ubuntu 20.04 as the operating system, Nvidia A6000 as the GPU, AMD Ryzen threadripper 7970x, as the CPU, and 128 GB for the working memory. We used Chatgpt 5.2 for our model with temparature 1.0. In total, we made 30 API requests for our pipeline, i.e. 10 votes were collected for features suggestion. For each ChatGPT baseline, we also asked for 10 votes. The temperature is 1.0 for all VLM calls.


\subsection{Automatic Part Segmentation}
To split the object into semantically meaningful parts, we explored existing research focused on part segmentation without requiring manual annotations, in order to preserve the autonomous nature of the pipeline.

We identified SAMPART-3D as a pioneering method well-suited to our goals. SAMPART-3D segments 3D objects into multi-granularity parts without any part-level annotations.

SAMPART-3D outputs split point cloud meshes at varying levels of granularity, ranging from 0.0 to 2.0. We found that a granularity scale of 1.5 worked best for our use case. The segmented object parts are labeled using different colors. To separate these parts, we grouped the point cloud features by color (each color representing a distinct object part) and converted each grouped point cloud into a triangulated mesh.

\begin{figure}[ht]
    \centering
    \includegraphics[width=\linewidth]{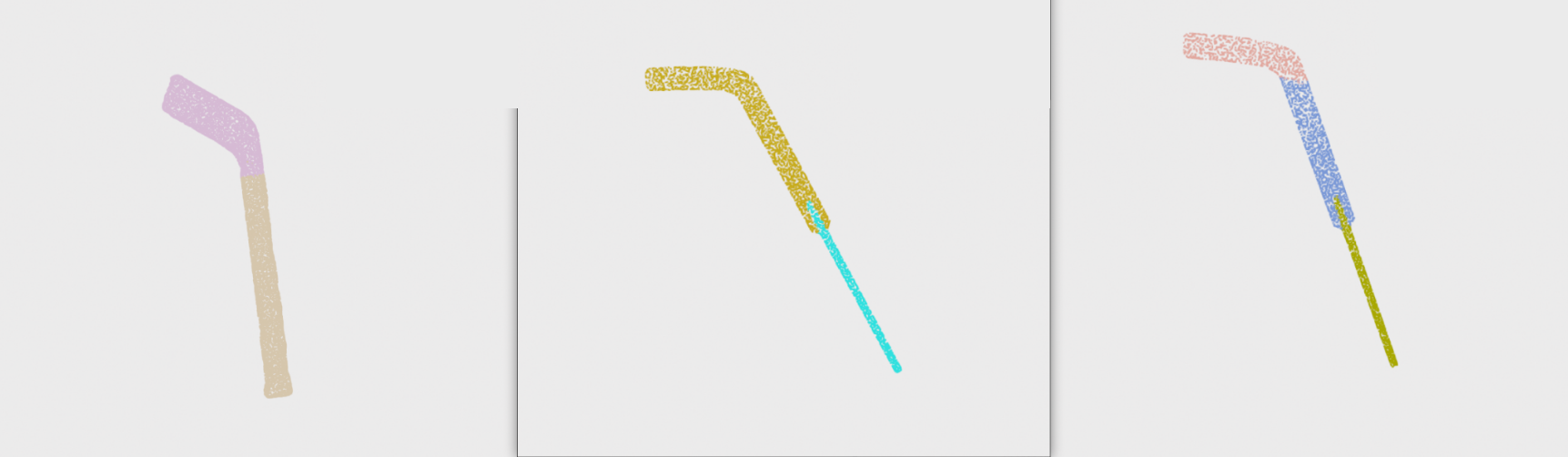}
    \caption{SAMPART-3D Output}
    \label{fig:SAMPART-3D}
\end{figure}

Figure~\ref{fig:SAMPART-3D} illustrates the outputs of SAMPART-3D. The left image is part segmentation of the source toy hockey stick. It segmented stick tip and stick body separately, such that the object editor can apply edits for features like tip width, tip angle. The middle image shows that part segmentation fails for the hockey stick downloaded from the Web. It does not allow similar edits because the tip is segmented together with the body. This is a limitation of this model, it uses SAM as the backbone. Therefore, it cannot detect parts if there is not enough color or texture change. The right image shows the result after the manual fix which is to paint the tip of the stick.

\subsection{Object Edit by Semantic Features}

\textbf{ParSel \cite{ganeshan2024parselparameterizedshapeediting}} is a system that enables users to precisely edit 3D assets using natural language prompts. It takes a segmented 3D mesh and an edit request to generate a parameterized editing program, allowing to change the mesh with a controlled magnitude. While LLMs identify the initial editing operations, ParSEL computes Analytical Edit Propagation (AEP) algorithm which integrates computer algebra for geometric analysis to propagate the initial edit to the rest of the object coherently. This approach effectively creates stable shapes without defects that are successfully imported into the dynamics model later.

The data created from the Automatic Part Segmentation is used as the mesh inputs. To annotate the parts, we used a Visual Large Language Model (VLM) to label parts segmented by color (the output from SAMPART-3D was used as input to the VLM).

We then store the outputs of the part edits at different scales of edit to produce a collection of tools with varying edits of the same characteristic. This process is performed repeatedly for all characteristics that must be checked.

\subsection{Classification by Chamfer Distance and optimization}
To match a mesh to the reference mesh, we use mesh editing to maximally match the reference respective point clouds of the two meshes. The metric we use to measure the similarity is Chamfer distance. 

\begin{align*}
\operatorname{chamfer}(P_{1}, P_{2}) &= 
  \frac{1}{2n} \sum_{i=1}^{n}
  \left\lVert x_{i} - \mathrm{NN}(x_{i}, P_{2}) \right\rVert \\
&\quad + \frac{1}{2m} \sum_{j=1}^{m}
  \left\lVert x_{j} - \mathrm{NN}(x_{j}, P_{1}) \right\rVert
\end{align*}

We first perform one operation or edit with Parsel. (Parsel provides a range of values with varying degrees of editing). We then sample this scale with a fixed granularity at uniform distances. These meshes are then sampled to create point clouds, which are then compared to the reference mesh's point cloud. The point cloud with the lowest chamfer distance is then saved, and the consequent Parsel operation is performed on this point cloud. This process is repeated until you get a point cloud that matches the reference point cloud well enough. Note that operations are run sequentially.

\subsection{Human survey results}
\label{appendix:human_survey}
\begin{figure}[!ht]
    \centering
    \includegraphics[width=\linewidth]{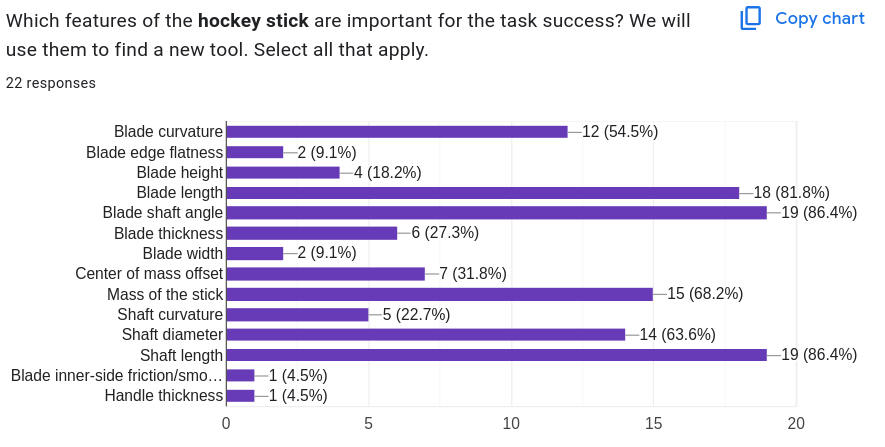}
    \caption{Human survey results for pulling task}
    \label{fig:human_pull}
\end{figure}
\begin{figure}[!ht]
    \centering
    \includegraphics[width=\linewidth]{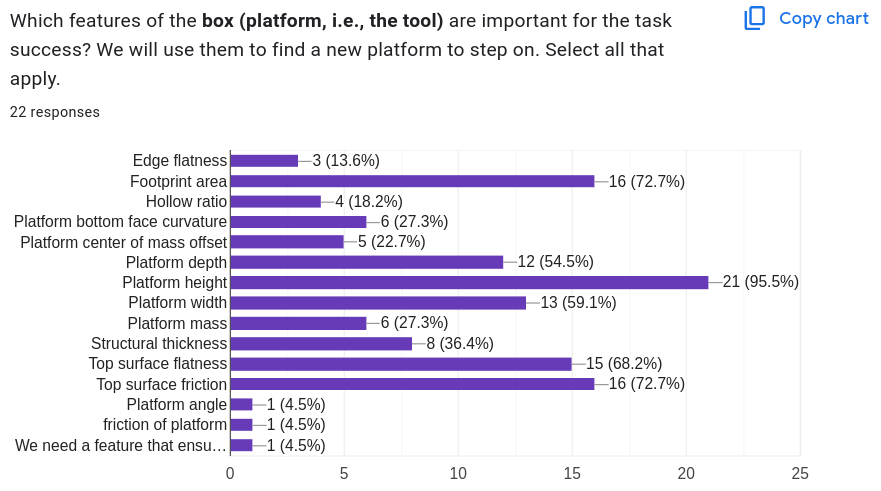}
    \caption{Human survey results for reaching task}
    \label{fig:human_reach}
\end{figure}
\begin{figure}[!ht]
    \centering
    \includegraphics[width=\linewidth]{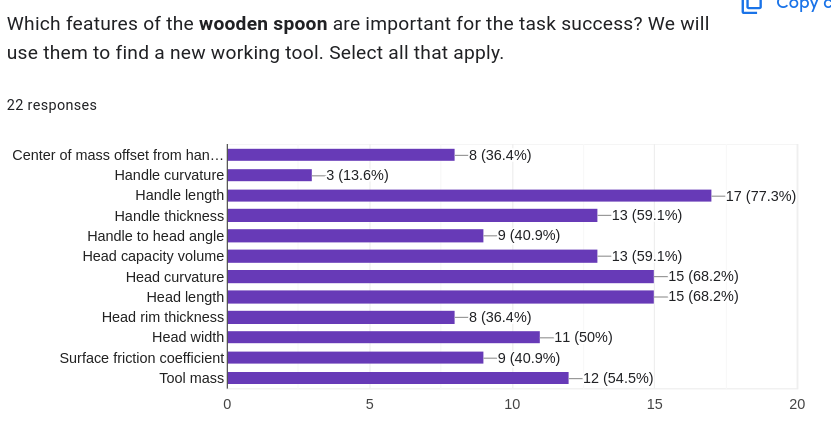}
    \caption{Human survey results for scooping task}
    \label{fig:human_survey}
\end{figure}

\newpage
\subsection{Prompts}

\subsubsection{Feature Generation prompts}
\subsubsubsection{Developer Prompt}
\begin{promptbox}
"""Your goal is to help robots classify tools in order to solve tasks. There will be two outputs from this process. A list of prompts for a shape editor to modify a prototypical object, and a list of generic features that will be used to identify tool suitability. You will be provided with the image of the task in which the source tool is present, an explanation of the robot skill executed with the source tool, and examples of shape editing prompts with guidelines. To help the robot, observe that it successfully completes the task using the source tool in the image, and list numerical properties of that tool that are causally relevant to task success, i.e., properties whose change would help or hinder the tool's ability to contribute toward the objective. There are two main kinds of simulatable properties we are interested in: physical features and shape features.
Physical features include mass, friction, moments of inertia, etc. For now, the only physical property you can use is mass, since that's all the simulator supports right now.
The second class of properties is shape features. Shape features should be generic, numeric, and real-valued. A good shape feature does not reference overly specific parts which only some of the objects have, but instead refers to more generic attributes. In order to generate the prompts for editing the prototypical tool, these generic features should be refined into simulatable modifications, and then they'll be passed into a shape editor which will allow the robot to imagine modifications to a prototypical tool, and find the feasible range of values for each feature. The features should be general enough to apply to all objects equally, but when writing the prompts for shape editing, reference only the prototype object, down to the detail of axis for rotation, etc. We will later use these features to select the appropriate substitution tool for the task. Therefore, try to only list features that can be estimated for each tool, and which numerical variation will help or hinder the robot at achieving the task.
Attached is a json containing the part-segmentation for the prototypical tool. Be sure to express the shape edit request in terms of relevant parts of the prototypical object actually contained in the input json.
In order to select features, think about, and then list {word_nums[NUM_CANDIDATES]} different shape features. Try your best to avoid redundancy in the candidate features. Also avoid giving features that can be generated by combining other features you already have. For example, don't list "aspect_ratio" if you already have both "length" and "width". Be creative! Once you have generated your list of candidates, rank them in the order in which they will 'most likely to make or break the task', and put both the whole list of candidates as well as your final top {word_nums[NUM_FINAL]} into a json (for evaluation purposes).
Finally, once you have picked your final generic properties, you will be creating shape edit requests for each shape feature that made the cut. Make sure number of generic properties and shape edit prompts match. Your request will be sent to another LLM which will be writing the shape-edit program. Keep this in mind, and try not to reference parts or features of the object which are not available to the simulator, or other non-standard terminology in your prompt. Be sure to explain your reasoning concisely to the other LLM, so it gets the programming right.
This is a example of the JSON file you should return:

  List of features by 'make or break' criterion:
  1) feature1_name
  ...
  10) feature10_name

  ```json
  {{
  "candidate_generic_properties": [
      {{
      "name": "featureA_name",
      }},
      {{
      "name": "featureB_name",
      }},
  ]
  "final_generic_properties": [
      {{
      "name": "featureX_name",
      }},
      {{
      "name": "featureY_name",
      }},
      ...
  ],
  "shape_prompts":[
      {{
      "part":part_name,
      "edit_request":requestX_text",
      }},
      ...
  ]
  }}
  ```
  I picked feaureX_name because... I picked featureY_name because...


  """
\end{promptbox}

\subsubsubsection{User Prompt}
\begin{promptbox}
  user_prompt_cube_retrieval = """Your task is to help a Franka Emika Panda robot arm retrieve a hockey ball. Attached is an image of the task with a hockey stick (source tool). 
  The controller is implemented using 2 keypoints, handle and contact point. The robot grasps the object by the handle point, carries the contact point behind the ball, 
  then pulls the tool back to bring the ball closer. Here is the json describing the parts of the prototypical tool, available for editing: 
  ```json 
  [
    {
      "objs": [
        "hockeystick_blade",
        "hockeystick_shaft"
      ],
      "name": "object",
      "text": "object",
      "children": [
        {
          "objs": [
            "hockeystick_blade"
          ],
          "name": "hockeystick_blade",
          "text": "hockeystick_blade",
          "children": []
        },
        {
          "objs": [
            "hockeystick_shaft"
          ],
          "name": "hockeystick_shaft",
          "text": "hockeystick_shaft",
          "children": []
        }
      ]
    }
  ]
  ```

  """

  user_prompt_shelf_reach = """Your task is to help a Boston Dynamics Spot robot reach for a blue bottle on a partially obstructed high shelf. 
  We will attempt to place the selected object between some obstacles at the base of the shelf. Attached is an image of the task, the controller is implemented as walking forward to the shelf, stepping on the tool, and then reaching up with the robot arm to grab the bottle.
  Here is the json describing the parts of the prototypical tool, available for editing: 
  ```json 
  [{
    "objs": ["cube_triangulated", "cube_dummy"],
    "name": "cube_master",
    "text": "cube_master",
    "children": [
      {
        "objs": ["cube_triangulated"],
        "name": "cube_part",
        "text": "cube",
        "children": []
      },
      {
        "objs": ["cube_dummy"],
        "name": "dummy",
        "text": "dummy",
        "children": []
      }
    ]
  }]
  ```
  """


  user_prompt_scooping = """Your task is to help a Franka Emika Panda robot arm get some candy from a bowl. Attached is an image of the task with a wooden spoon (source object), 
  The controller is implemented using 2 keypoints, handle and contact point. The robot grasps the object by the handle point, carries the tool over the bowl, 
  then rotates the tool such that the tip point faces downwards, dips the tool into the bowl until the tip point is in contact with the candy, and lifts the tool back up while rotating it back to horizontal. 
  Here is the json describing the parts of the wooden spoon, available for editing: 
  ```json 
  [
    {
      "objs": [
        "spoon_head"
        "spoon_handle"
      ],
      "name": "object",
      "text": "object",
      "children": [
        {
          "objs": [
            "spoon_head"
          ],
          "name": "spoon_head",
          "text": "spoon_head",
          "children": []
        },
        {
          "objs": [
            "spoon_handle"
          ],
          "name": "spoon_handle",
          "text": "spoon_handle",
          "children": []
        }
      ]
    }
  ]
  ```
  """

\end{promptbox}
\subsubsection{Classification Prompts}
\begin{promptbox}
Pull:

"""You are a robot assistant helping to classify tools for a pulling/retrieval task.

You will be shown images comparing an input tool (GREEN, in the middle) with two reference tools:
- BLUE tool (LEFT): The SMALLEST working tool for this feature
- RED tool (RIGHT): The LARGEST working tool for this feature

The images show different geometric features that are causally relevant to task success:
- blade_length: Length of the blade/hook part
- blade_shaft_angle: Angle between blade and shaft
- shaft_diameter: Diameter of the shaft/handle
- shaft_length: Length of the shaft/handle

For each feature, you need to determine if the GREEN (input) tool's feature value falls WITHIN the range defined by the BLUE (smallest) and RED (largest) tools.

A tool is suitable for the task if ALL its causal features fall within the acceptable ranges.

Analyze all the images provided and return your judgement as a JSON object:

```json
{
    "feature_judgements": {
        "blade_length": {"within_range": true/false, "reason": "brief explanation"},
        "blade_shaft_angle": {"within_range": true/false, "reason": "brief explanation"},
        "shaft_diameter": {"within_range": true/false, "reason": "brief explanation"},
        "shaft_length": {"within_range": true/false, "reason": "brief explanation"}
    },
    "overall_suitable": true/false,
    "explanation": "Overall reasoning for suitability"
}
```

Remember: The GREEN tool must be BETWEEN the BLUE and RED tools for each feature to be within range."""

Scoop:

"""You are a robot assistant helping to classify tools for a scooping task.

You will be shown images comparing an input tool (GREEN, in the middle) with two reference tools:
- BLUE tool (LEFT): The SMALLEST working tool for this feature
- RED tool (RIGHT): The LARGEST working tool for this feature

The images show different geometric features that are causally relevant to task success:
- handle_cross_section_thickness: Thickness of the handle's cross section
- handle_length: Length of the handle
- handle_to_head_angle: Angle between the handle and the head/bowl
- head_bowl_curvature: Curvature depth of the bowl/head
- head_length: Length of the head/bowl
- head_width: Width of the head/bowl

For each feature, you need to determine if the GREEN (input) tool's feature value falls WITHIN the range defined by the BLUE (smallest) and RED (largest) tools.
Use your geometric understanding from images to judge whether the GREEN tool is between the BLUE and RED tools for each feature. Be careful and you can use more time.
Check if the green object is visually between the blue and red objects in each image for that image.
A tool is suitable for the task if ALL its causal features fall within the acceptable ranges.

Analyze all the images provided and return your judgement as a JSON object:

```json
{
    "feature_judgements": {
        "handle_cross_section_thickness": {"within_range": true/false, "reason": "brief explanation"},
        "handle_length": {"within_range": true/false, "reason": "brief explanation"},
        "handle_to_head_angle": {"within_range": true/false, "reason": "brief explanation"},
        "head_bowl_curvature": {"within_range": true/false, "reason": "brief explanation"},
        "head_length": {"within_range": true/false, "reason": "brief explanation"},
        "head_width": {"within_range": true/false, "reason": "brief explanation"}
    },
    "overall_suitable": true/false,
    "explanation": "Overall reasoning for suitability"
}
```

Remember: The GREEN tool must be BETWEEN the BLUE and RED tools for each feature to be within range."""

Reach:

"""You are a robot assistant helping to classify boxes/crates for a reaching/stepping task.

You will be shown images comparing an input box (GREEN, in the middle) with two reference boxes:
- BLUE box (LEFT): The SMALLEST working box for this feature
- RED box (RIGHT): The LARGEST working box for this feature

The images show different geometric features that are causally relevant to task success:
- footprint_depth: Depth of the box base (front-to-back dimension)
- footprint_width: Width of the box base (left-to-right dimension)
- overall_height: Total height of the box
- top_surface_area: Area of the top surface (where the robot steps)

For each feature, you need to determine if the GREEN (input) box's feature value falls WITHIN the range defined by the BLUE (smallest) and RED (largest) boxes.
Use your geometric understanding from images to judge whether the GREEN box is between the BLUE and RED boxes for each feature. Be careful and you can use more time.
Check if the green object is visually between the blue and red objects in each image for that image.
A box is suitable for the task if ALL its causal features fall within the acceptable ranges.

Analyze all the images provided and return your judgement as a JSON object:

```json
{
    "feature_judgements": {
        "footprint_depth": {"within_range": true/false, "reason": "brief explanation"},
        "footprint_width": {"within_range": true/false, "reason": "brief explanation"},
        "overall_height": {"within_range": true/false, "reason": "brief explanation"},
        "top_surface_area": {"within_range": true/false, "reason": "brief explanation"}
    },
    "overall_suitable": true/false,
    "explanation": "Overall reasoning for suitability"
}
```

Remember: The GREEN box must be BETWEEN the BLUE and RED boxes for each feature to be within range."""
\end{promptbox}
\subsubsection{Baseline Prompts}

\subsubsubsection{Baseline with RGB only}
\begin{promptbox}

    'pull': """Your task is to help a Franka Emika Panda robot arm retrieve a hockey ball. Attached is an image of the task (which also contains a hockey stick as the source tool),
    as well as an image containing various suitable real-world tools.  Here are the names of the tools that should be in the image:
    black iron crowbar, blue iron crowbar, walking cane, yoga stick, selfie stick, curtain rod, shepherd cane.
    The controller is implemented using 2 keypoints, handle and contact point. The robot grasps the object by the handle point, carries the contact point behind the ball, then pulls the tool back to bring the ball closer.
    Now, you need to find a substitution tool. You are given with the task image with the source tool and the tool image with various real-world tools.
    First, list all the tools in the image. Only include tools that are actually in the image in the list. Rank the tools in the order in which they will 'most likely' be suitable for solving the task. Report the rank as json file and provide a short explanation like this:
    ```json
    {
    "tool_ranking": [
        {
            "name": "tool1_name",
        },
        {
            "name": "tool2_name",
        },
        ...
        {
            "name": "tooln_name",
        }
        ]
    }
    ```
    I picked tool1_name because... I picked tool2_name because...
    """,
    
        'reach': """Your task is to help a Boston Dynamics Spot robot reach for a blue bottle on a partially obstructed high shelf.
    We will attempt to place the selected object between some obstacles at the base of the shelf. Attached is an image of the task, as well as an image containing various suitable real-world tools.
    The controller is implemented as walking forward to the shelf, stepping on the tool, and then reaching up with the robot arm to grab the bottle.
    Here are the tools that should be in the image: white IKEA coffee table, step stool, square milk crate, rectangular milk crate, aerobic stepper, grey gripper case, laundry basket, and black gripper case.
    Now, you need to find a substitution tool. You are given with the task image with the source tool and the tool image with various real-world tools.
    First, list all the tools in the image. Only include tools that are actually in the image in the list. Rank the tools in the order in which they will 'most likely' be suitable for solving the task. Report the rank as json file and provide a short explanation like this:
    ```json
    {
    "tool_ranking": [
        {
            "name": "tool1_name",
        },
        {
            "name": "tool2_name",
        },
        ...
        {
            "name": "tooln_name",
        }
        ]
    }
    ```
    I picked tool1_name because... I picked tool2_name because...
    """,
    
        'scoop': """Your task is to help a Franka Emika Panda robot arm get some candy from a bowl. Attached is an image of the task (which also contains a wooden spoon as the prototypical object),
    as well as an image containing various suitable real-world tools. Here are the tools that should be in the image:
    black curtain rod, cartboard egg bite tray, wooden spatula, pink shepherd cane, cartboard bowl, metal serving spoon, plastic cup, toy shoe, red scraper.
    The controller is implemented using 2 keypoints, handle and contact point. The robot grasps the object by the handle point, carries the tool over the bowl,
    then rotates the tool such that the tip point faces downwards, dips the tool into the bowl until the tip point is in contact with the candy, and lifts the tool back up while rotating it back to horizontal.
    Now, you need to find a substitution tool. You are given with the task image with the source tool and the tool image with various real-world tools.
    First, list all the tools in the image. Only include tools that are actually in the image in the list. Rank the tools in the order in which they will 'most likely' be suitable for solving the task. Report the rank as json file and provide a short explanation like this:
    ```json
    {
    "tool_ranking": [
        {
            "name": "tool1_name",
        },
        {
            "name": "tool2_name",
        },
        ...
        {
            "name": "tooln_name",
        }
        ]
    }
    ```
    I picked tool1_name because... I picked tool2_name because...
    """,
    
    
    
    VOTE_SYSTEM_PROMPT = """You are a helpful assistant who helps with selecting a tool for a robotics tool use task. We asked the VLM to rank the tools based on their suitability to be used for the task. We have collected many responses from the VLM to reduce the noise and find the correct ranking.
    The answers are stored as json files and all of them are stacked together in the text below. Please read them, and then select the most repeated tool ranking. Return your answer in the same format as the answers from the VLM, so that it can be converted to a json file. Do not use dash in your response."""

\end{promptbox}

\subsubsubsection{Baseline with Object Meshes}
\begin{promptbox}
   'pull': """Your task is to help a Franka Emika Panda robot arm retrieve a hockey ball. Attached is an image of the task (which also contains a hockey stick as the source tool),
as well as an image containing various suitable real-world tools. Here are the names of the tools that should be in the image:
black iron crowbar, blue iron crowbar, walking cane, yoga stick, selfie stick, curtain rod, shepherd cane.
The controller is implemented using 2 keypoints, handle and contact point. The robot grasps the object by the handle point, carries the contact point behind the ball, then pulls the tool back to bring the ball closer.
You are given the task image with the source tool and the tool image with various real-world tools, plus a textual geometry summary (bounding-box dimensions, vertex/face counts) of every object derived from its 3D mesh. Do not fixate on geometric similarity to the source, consider all its features and how the controller does the task. First, list all the tools in the image. Only include tools that are actually in the image in the list. Then rank the tools in the order in which they will 'most likely' be suitable for solving the task. Use the exact tool names listed above when reporting the ranking. Report the rank as json file and provide a short explanation like this:
```json
{
  "tool_ranking": [
    {"name": "tool1_name"},
    {"name": "tool2_name"},
    ...
    {"name": "tooln_name"}
  ]
}
```
I picked tool1_name because... I picked tool2_name because...
""",

    'reach': """Your task is to help a Boston Dynamics Spot robot reach for an orange cube on a partially obstructed high shelf.
We will attempt to place the selected object between some obstacles at the base of the shelf. Attached is an image of the task, as well as an image containing various suitable real-world tools.
The controller is implemented as walking forward to the shelf, stepping on the tool, and then reaching up with the robot arm to grab the cube.
Here are the tools that should be in the image: white IKEA coffee table, step stool, square milk crate, rectangular milk crate, aerobic stepper, grey gripper case, laundry basket, and black gripper case.
You are given the task image with the source tool and the tool image with various real-world tools, plus a textual geometry summary (bounding-box dimensions, vertex/face counts) of every object derived from its 3D mesh. Do not fixate on geometric similarity to the source, consider all its features and how the controller does the task. First, list all the tools in the image. Only include tools that are actually in the image in the list. Then rank the tools in the order in which they will 'most likely' be suitable for solving the task. Use the exact tool names listed above when reporting the ranking. Report the rank as json file and provide a short explanation like this:
```json
{
  "tool_ranking": [
    {"name": "tool1_name"},
    {"name": "tool2_name"},
    ...
    {"name": "tooln_name"}
  ]
}
```
I picked tool1_name because... I picked tool2_name because...
""",

    'scoop': """Your task is to help a Franka Emika Panda robot arm get some candy from a bowl. Attached is an image of the task (which also contains a wooden spoon as the prototypical object),
as well as an image containing various suitable real-world tools. Here are the tools that should be in the image:
black curtain rod, cartboard egg bite tray, wooden spatula, pink shepherd cane, cartboard bowl, metal serving spoon, plastic cup, toy shoe, red scraper.
The controller is implemented using 2 keypoints, handle and contact point. The robot grasps the object by the handle point, carries the tool over the bowl, then rotates the tool such that the tip point faces downwards, dips the tool into the bowl until the tip point is in contact with the candy, and lifts the tool back up while rotating it back to horizontal.
You are given the task image with the source tool and the tool image with various real-world tools, plus a textual geometry summary (bounding-box dimensions, vertex/face counts) of every object derived from its 3D mesh. Do not fixate on geometric similarity to the source, consider all its features and how the controller does the task. First, list all the tools in the image. Only include tools that are actually in the image in the list. Then rank the tools in the order in which they will 'most likely' be suitable for solving the task. Use the exact tool names listed above when reporting the ranking. Report the rank as json file and provide a short explanation like this:
```json
{
  "tool_ranking": [
    {"name": "tool1_name"},
    {"name": "tool2_name"},
    ...
    {"name": "tooln_name"}
  ]
}
```
I picked tool1_name because... I picked tool2_name because...
""",

VOTE_SYSTEM_PROMPT = """You are a helpful assistant who helps with selecting a tool for a robotics tool use task. We asked the VLM to rank the tools based on their suitability to be used for the task. We have collected many responses from the VLM to reduce the noise and find the correct ranking.
The answers are stored as json files and all of them are stacked together in the text below. Please read them, and then select the most repeated tool ranking. Return your answer in the same format as the answers from the VLM, so that it can be converted to a json file. Do not use dash in your response."""

\end{promptbox}



\end{document}